\documentclass[review]{elsarticle}

\usepackage{times}
\usepackage{graphicx}
\usepackage{amsmath}
\usepackage{amssymb}
\usepackage{algorithm}
\usepackage{algorithmicx}
\usepackage{algpseudocode}
\usepackage{multirow}
\usepackage{color} 
\usepackage{subfigure}
\usepackage{float}

\journal{Journal of \LaTeX\ Templates}

\usepackage{xspace}

\makeatletter
\DeclareRobustCommand\onedot{\futurelet\@let@token\@onedot}
\def\@onedot{\ifx\@let@token.\else.\null\fi\xspace}

\def\eg{\emph{e.g}\onedot} 
\def\ie{\emph{i.e}\onedot} 
 
\def\etc{\emph{etc}\onedot} 
 
\def\etal{\emph{et al}\onedot}
\makeatother

\begin{document}

\begin{frontmatter}

\title{Learning Fine-grained Features via a CNN
Tree for Large-scale Classification}

\author[mymainaddress]{Zhenhua Wang\corref{mycorrespondingauthor}}
\cortext[mycorrespondingauthor]{Corresponding author}
\ead{zhenhua.wang@mail.huji.ac.il}

\author[mysecondaryaddress]{Xingxing Wang\textsuperscript{+,}\footnotetext[1]{\textsuperscript{+}Joint first author} }
\author[mythirdaddress]{Gang Wang}

\address[mymainaddress]{School of Computer Science and Engineering,The Hebrew University of Jerusalem, Jerusalem, Israel}
\address[mysecondaryaddress]{School of Electrical and Electronic Engineering, Nanyang Technological University, Singapore }
\address[mythirdaddress]{Alibaba AI Labs, Hangzhou, China}

\begin{abstract}
We propose a novel approach to enhance
the discriminability of Convolutional Neural Networks (CNN).
The key idea
is to build a tree structure that could
progressively learn fine-grained
features to distinguish a subset of classes, by learning features 
only among these classes. Such features 
are expected to be more discriminative, compared to features learned for all the
classes. We develop a new algorithm to effectively
learn the tree structure from a large number of classes.
Experiments on large-scale image classification tasks
demonstrate that our method could boost the performance of a given basic CNN model.
Our method is quite general, hence it can potentially be used 
in combination with many other deep learning models.
\end{abstract}


\end{frontmatter}

\section{Introduction\label{sec:intro}}

Convolutional neural networks (CNN)~\cite{LeCun_back_1989,lecun1998gradient} have recently demonstrated superior performance on many tasks such as 
image classification~\cite{Alex_imagenet_2012,Zeiler_visual_2014,He_delving_2015},
object detection~\cite{Girshick_rich_2014,Girshick_fast_2015,Pierre_overfeat_2013,ouyang2015deepid, Szegedy_scalable_2014,erhan2014scalable}, 
object tracking~\cite{li2015robust,wang2015video,hong2015online}, 
text detection~\cite{delakis2008text,huang2014robust}, 
text recognition~\cite{Goodfellow_2013_Multi,jaderberg2014deep,wang2012end}, local feature description~\cite{Tian_2017},
video classification~\cite{simonyan2014two,karpathy2014large,ng2015beyond}, 
human pose estimation~\cite{toshev2014deeppose,jain2013learning,tompson2014joint},
scene recognition~\cite{Gong_multi_2011,Zhou_learn_2014} and 
scene labelling~\cite{shuai2015integrating,farabet2013learning}. 
In the classical CNN framework ~\cite{Alex_imagenet_2012}, 
a CNN learns 7 shared feature layers (5 convolutional layers and 2 fully-connected layers) for all
the categories, followed by the last fully-connected layer to distinguish among different classes based on the learned features. 
Such learned features encode  discriminative visual patterns, which are useful for classification. 
However, for different subsets of categories, discriminative visual patterns should vary. For example, features used to distinguish cat and dog should be different from those used to distinguish grass and tree. In the current CNN model, we just fuse different types of discriminative features together in a single model for different subsets of categories. 
Then for a certain subset of classes, features learned to distinguish other classes might become noise. Hence, such a single CNN model may not be
ideal for classification, due to the lack of enough representation capacity. 

In this paper, we present a novel approach
to enhance the discriminability of CNN for large-scale multiclass classification.
The key idea is to learn fine-grained features specifically for a subset of classes.
Compared to features learned for all the categories, 
such fine-grained features are expected to better capture discriminative visual patterns
of the subset of classes.
Ideally, for a class $c_i$, if we can identify the confusion set $\mathcal{S}_i$ which contains all the possible classes whose testing examples might be classified as $c_i$ by the basic CNN model, then we can learn a specific CNN model for this confusion set $\mathcal{S}_i$ only. 
Since $\mathcal{S}_i$ contains fewer categories than the whole class set, the specific CNN model should encode discriminative visual patterns which are more useful for this confusion set.
Then testing examples which are misclassified as $c_i$ by the basic CNN model might be corrected
by the specific CNN model.

\begin{figure*}[t]
\begin{center}
\includegraphics[width=1.0\linewidth]{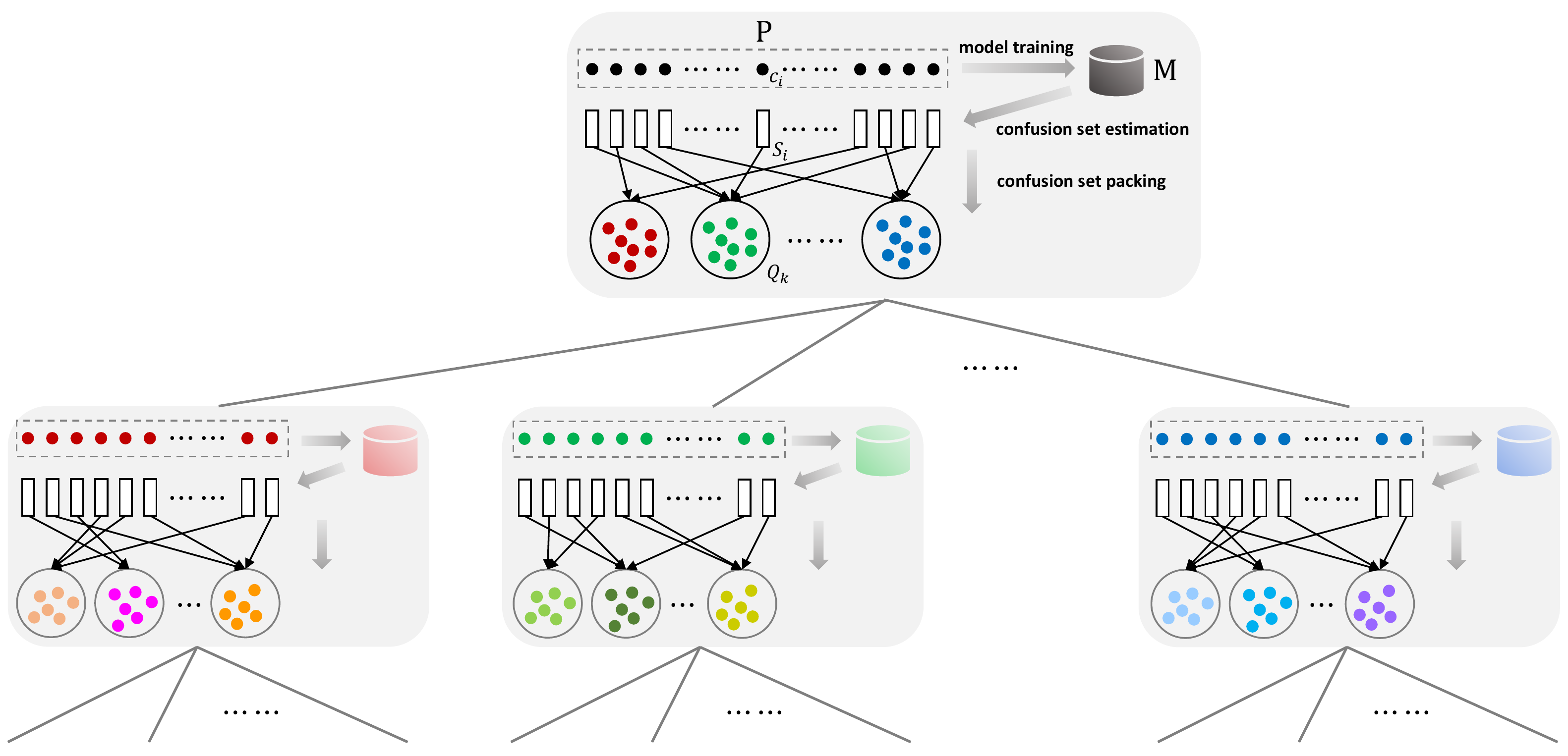}
\end{center}
   \caption{The learning procedure of a CNN tree. 
Given a node of the tree, we first train a model on its class set.
Next we estimate the confusion set of each class by using the trained model.
Then these confusion sets are packed into several confusion supersets and 
each of them is assigned to a new child node for further learning. This procedure repeats until it reaches the maximal depth.}
\label{fig:cnn_tree}
\end{figure*}

However, when there are a large number of categories (\eg, 1000), 
it is computationally expensive to learn a specific CNN model for each class. 
And worse, in this case, it might be difficult to cache all the CNN models in the memory in the testing procedure. 
Then the testing speed will be significantly decreased as we 
have to frequently load models from the hard disk. 
Hence we need to develop advanced methods to reduce the number of specific CNN models to train.

We observe that some classes share many common categories in their confusion sets. 
Then we can train a shared CNN model for these classes by merging their confusion sets.
We may sacrifice accuracy because now our specific CNN model 
deals with more categories, 
 but at the same time, we gain on reducing the number of specific CNN models to train. 
 We can iteratively repeat this procedure: 
 given the new CNN model, we re-estimate the confusion set for each class in that model, 
 and train a specific CNN model for each confusion set respectively afterwards.
 At the end, we will build a tree of CNN models, 
 each of which aims to learn discriminative features for a number of categories only. 
 
 When we merge the confusion sets, we should make a trade-off 
 between efficiency and accuracy.
 We pose the optimization problem as minimizing the number of merged confusion sets while 
 requiring the error rate of resulting CNN model to be below a certain threshold.
 This objective is intractable.
 Thus, we relax the constraint on accuracy by limiting the number of categories in each merged set.   
 This assignment problem is similar to the virtual machine packing problem and is an NP-hard problem. 
 And proably good polytime approximation algorithms may not exist~\cite{sindelar_sharing-aware_2011}.
In this paper, a heuristic algorithm is then proposed based on the intuition that
confusion sets with more overlap should be merged together in higher priority in order to optimize the
benefits of inter-set category sharing.

To evaluate the proposed method, we use both AlexNet~\cite{Alex_imagenet_2012} and GoogleNet~\cite{Christian_going} as our basic CNN models.
Experiments on the ILSVRC 2015 dataset~\cite{ILSVRC_2015} show that our method could improve the performance of both models.
This sufficiently demonstrates the effectiveness of the proposed method.
Our method is quite general, hence it can potentially be used in combination with many other deep learning models.

The rest of the paper is organized as follows: Section~\ref{sec:related} gives an overview of related work.
Section~\ref{sec:tree} introduces a novel CNN tree and gives a detailed description of the tree learning algorithm.
The experimental evaluations are carried out
in section~\ref{sec:exp} and finally we conclude the paper in section~\ref{sec:conclusion}.

\begin{table}
\begin{center}
\begin{tabular}{|c|c|}
\hline
Notation &  Description  \\
\hline\hline
$T$ &  a CNN tree\\
\hline
$v$ & a node of a CNN tree  \\
\hline
$\phi(v)$ & the children nodes set of node $v$\\
\hline
$\mathcal{P}(v) $ &  the class set of node $v$\\
\hline
$M(v)$ &  the CNN model trained on $\mathcal{P}(v)$\\
\hline
$c_i$ & the $i^{th}$ class of a class set\\
\hline
$\mathcal{S}_i$ & the confusion set of $c_i$\\
\hline
$\mathcal{Q}_k$ &  the $k^{th}$ confusion superset\\
\hline
\end{tabular}
\end{center}
\caption{The notations for the CNN Tree.}
\label{tb:notation}
\end{table}

\section{Related Work\label{sec:related}}

Convolutional Neural Networks have a long history in computer vision.
Since its introduction by LeCun \etal \cite{LeCun_back_1989}, 
it has consistently been competitive with other methods for recognition tasks.
Recently, with the advent of large-scale category-level training data, \eg ImageNet~\cite{deng_imagenet_2009}, 
CNN exhibit superior performance in large-scale visual recognition.
Most notably, Krizhevsky \etal \cite{Alex_imagenet_2012} proposed a classic CNN architecture which 
contains eight learned layers (5 convolutional layers and 3 fully-connected layers), and showed 
significant improvement upon previous methods on the image classification task.

Several techniques have been proposed to boost the performance of CNN in different aspects.
Hinton~\etal~\cite{Hinton_dropout} proposed a method called Dropout to prevent complex co-adaptation on the training data.
This method simply omits half of the feature detectors on each training case, so a hidden unit cannot rely on other hidden units 
being present. Wan~\etal~\cite{wan2013regularization} introduced DropConnect which is a generation of Dropout for regularizing 
large fully-connected layers within neural networks.
Unlike Dropout, DropConnect sets a randomly selected subset of weights within the network to zero. Each unit thus receives input from a random subset of units in the previous layer.
Ciresan~\etal~\cite{Ciresan_multi_column} proposed a multi-column CNN which improved the performance by combing several CNN columns together.
Goodfellow~\etal~\cite{goodfellow2013maxout} proposed a new model called Maxout which aims to both facilitate optimization by Dropout and improve the accuracy of Dropout's fast approximate model average technique.
Howard~\cite{Andrew_improve_2013} presented some new useful image transformation to increase the effective size of training set
and generate more test predictions. They also showed an efficient way to train higher resolution models that generate useful complementary predictions.
Wu~\etal~\cite{wu2015deep} also proposed more complex data augmentation techniques for model training.
He~\etal~\cite{Kaiming_spatial} proposed a new network structure called SPP-net which
equipped the CNN with a more principled pooling strategy, \ie "spatial pyramid pooling".
The SPP-net can generate a fixed-length representation regardless of the image size. Also, it is robust 
to object deformations due to the use of pyramid pooling.
Ioffe~\etal~\cite{ioffe2015batch} proposed a method called Batch Normalization which reduced the internal covariate shift by
normalizing each training mini-batch. This method allows the training procedure to use
much higher learning rate and be less careful about initialization, and in some cases eliminates the needs for Dropout.
Zeiler and Fergus~\cite{Zeiler_visual} presented a novel visualization method that gives insight into the function of intermediate feature layers and the operation of the classifier, and showed how these visualization can be used to identify problems of
the model and so obtain better performance. Simply by reducing the layer filter size and the stride of convolution, 
they significantly improved the performance.
Simonyan and Zisserman~\cite{simonyan2014very} investigated the effect of CNN depth on its accuracy in large-scale image classification. They showed the the representation depth is beneficial for the classification
accuracy.
Szegedy~\etal~\cite{Christian_going} proposed a new CNN architecture
which achieved high performance by increasing the depth and width of the network.
He~\etal~\cite{he2016deep} introduced a residual learning framework to ease the training of networks.
Their networks are much deeper than all existing works, and achieved the-state-of-the-art performance.
All of these methods learn generic features from the whole class set, 
which might not be discriminative for classes in a specific subset.
Differently, our method progressively learns fine-grained features 
for each class subset in the tree. These features should be more discriminative in distinguishing a specific class subset
than the generic features learned on the whole class set.

Another kind of methods that is related to our work is the hierarchical classification.
A number of papers have proposed to exploit the hierarchical structure between object classes.
Griffin and Perona~\cite{griffin_learning_2008} propose an algorithm for automatically building classification trees.
Deng \etal \cite{Deng_what_2010} and Torralba~\etal \cite{ Torralba_2008} 
propose to exploit semantic hierarchy from WordNet to improve classification.
Bengio \etal \cite{bengio_label_2010} propose a label embedding tree for large-scale classification.
It partitions categories into disjoint subsets recursively, and train one classifier for each subset.
Deng \etal \cite{deng_fast_2011} improve the label tree
by simultaneously determining the tree structure and learning the classifier 
for each node in the tree. They also allow overlaps among different subsets.
The main purpose of these hierarchical methods is to speed up the operation of testing.
However, test examples that are misclassified at top layers can not be 
recovered later in their methods.
In contrast, our method is able to recover test examples misclassified by the basic model 
as it can learn fine-grained features to distinguish the confusion set.

\section{The Proposed CNN Tree\label{sec:tree}}

\subsection{Overview\label{sec:overview}}
We are motivated by the observation that a class is usually confused by a few number of other classes
in multiclass classification. 
In this paper, we refer to them as the confusion set of that class.
Given the confusion set of one class, more discriminative features could
be learned by a specific CNN to distinguish the classes only in this set.
Based on this observation, we develop a method to
 progressively learn fine-grained features for different confusion sets.

Suppose $\tilde{\mathcal{C}}$ is the whole class set of
a multiclass classification task and $\tilde{M}$ is a basic
CNN model learned on $\tilde{\mathcal{C}}$.
The confusion set $\mathcal{S}_i$ of a class $c_i$ 
is a set that contains $c_i$ and 
all other classes whose test examples are potential to be misclassified as $c_i$ by $\tilde{M}$.
With a large number of classes, we claim
$\left|\mathcal{S}_i\right| \ll |\tilde{\mathcal{C}}|$ 
based on the intuition that a class should only be confused by a 
few similar classes in the whole class set.
This exhibits valuable information for further improving the accuracy of $\tilde{M}$.
Specifically, we can train a specific CNN model
to learn fine-grained features for each $\mathcal{S}_i$ only.
Since $\mathcal{S}_i$ contains fewer classes, 
these fine-grained features should be more discriminative than the generic 
features learned by $\tilde{M}$ on the $\tilde{\mathcal{C}}$.
When a test example is classified as $c_i$ by $\tilde{M}$, 
we can then use the specific CNN model trained on $\mathcal{S}_i$ to refine its class label. It is hopeful that a test example that is misclassified by the basic CNN model might be correctly classified by the new specific CNN model.

However, when dealing with a large number of categories,
learning a specific CNN model for each class separately is very computationally expensive.
It is also difficult to cache all the CNN models in the
memory in the testing procedure, leading 
to a significant decrease of testing speed.
To alleviate this problem, we adopt a tree structure to progressively 
learn fine-grained features, namely the CNN tree.

Before we formally introduce the proposed CNN tree, 
we summarize the notations in Table~\ref{tb:notation} for clarity.
Let $T=(V,E)$ be a CNN tree. 
As shown in Fig.~\ref{fig:cnn_tree}, for a node $v\in V$, we are given a class subset $\mathcal{P}(v)\subset \tilde{\mathcal{C}}$.
We try to classify different classes in $\mathcal{P}(v)$ by learning a specific CNN model $M(v)$ on $\mathcal{P}(v)$.
If $v$ is not a leaf node, we need to further 
enhance the discriminability of $M(v)$.
To this end, we re-estimate the confusion set $\mathcal{S}_i$ of 
each $c_i\in \mathcal{P}(v)$ based on $M(v)$ using the method described in section~\ref{sec:confuse_estimation}.
Based on the observation that some classes may share many common categories in their confusion sets,
we can pack them into a confusion superset, \ie take the union of them, to reduce the number of CNNs to train.
During the confusion set packing procedure, one confusion set could only be packed to one superset.
Suppose we get $N$ confusion supersets after merging, 
$N$ children nodes will be generated respectively.
Let $\phi(v)\subset{V}$ be the children nodes set of node $v$.
For a child node $u_k\in\phi(v)$, the confusion superset $\mathcal{Q}_k$ is assigned
to the class subset $\mathcal{P}(u_k)$ for further learning specific CNN model on $\mathcal{Q}_k$. 

To classify an input instance $x$ with the constructed CNN tree, 
we use the prediction algorithms shown in Algorithm~\ref{alg:cnn_tree_predict}.
Specifically, we start at the root node $\tilde{v}$ by predicting the class label $c_l$ of $x$ with $M(\tilde{v})$.
To determine the child node for refining the prediction, 
we obtain the confusion set $\mathcal{S}_l$ of $c_l$ and
find the confusion superset $\mathcal{Q}_k$ which $\mathcal{S}_l$ is merged into.
Then, we go to child node $u_k$ where $\mathcal{P}(u_k) = \mathcal{Q}_k$.
Note that the selected child node is unique because 
the confusion set $\mathcal{S}_l$ is uniquely merged into one confusion superset during training.
At the child note $u_k$, we can use 
the specific CNN model $M(u_k)$ learned on $\mathcal{P}(u_k)$ to refine the prediction.  
This process repeats until it reaches a leaf node.
The class label predicted by the specific CNN model of this leaf node is regarded as the final result.

It is worth noting that the proposed method is quite different from 
the works on hierarchical classification~\cite{bengio_label_2010,deng_fast_2011,griffin_learning_2008}.
The main purpose of hierarchical classification is to speed up the operation of testing.
These methods partition categories into different subsets recursively, and train one classifier for each subset. However, test examples that are misclassified at top layers can not be recovered later in their methods. 
In contrast, our method refines the class labels predicted by the basic model 
via progressively learning more specific models on their confusion sets.
As the specific model could learn fine-grained features to distinguish confused classes,
it is capable of correcting test examples misclassified by the basic model.

\renewcommand{\algorithmicrequire}{\textbf{INPUT:}}
\renewcommand{\algorithmicensure}{\textbf{OUTPUT:}}
\newcommand{\algorithmicbreak}{\textbf{break}}
\newcommand{\Break}{\State \algorithmicbreak}

\begin{algorithm}
\begin{algorithmic}[1]
\Require The input instance $x$, the CNN tree $T$
\Ensure The predicted class label $c_l$.
\State $v \gets$ the root node of $T$ 
\While {$\phi(v) \neq \emptyset$}
\State Predict the class label $c_{l}$ of $x$ by using $M(v)$.
\State Obtain the confusion set $\mathcal{S}_{l}$, and find the confusion superset $\mathcal{Q}_k$ which $\mathcal{S}_{l}$ is merged into.
\State Fine the child node $u_k\in\phi(v)$ where $\mathcal{P}(u_k)=\mathcal{Q}_k$.
\State $v \gets u_k$
\EndWhile
\State Predict the class label $c_{l}$ of $x$ by using $M(v)$.
\end{algorithmic}
\caption{\emph{The prediction algorithm of a CNN tree}}
\label{alg:cnn_tree_predict}
\end{algorithm}

\subsection{Learning CNN Tree\label{sec:learning}}
\begin{figure*}[htbp] 
\centering\includegraphics[width=1.0\textwidth]{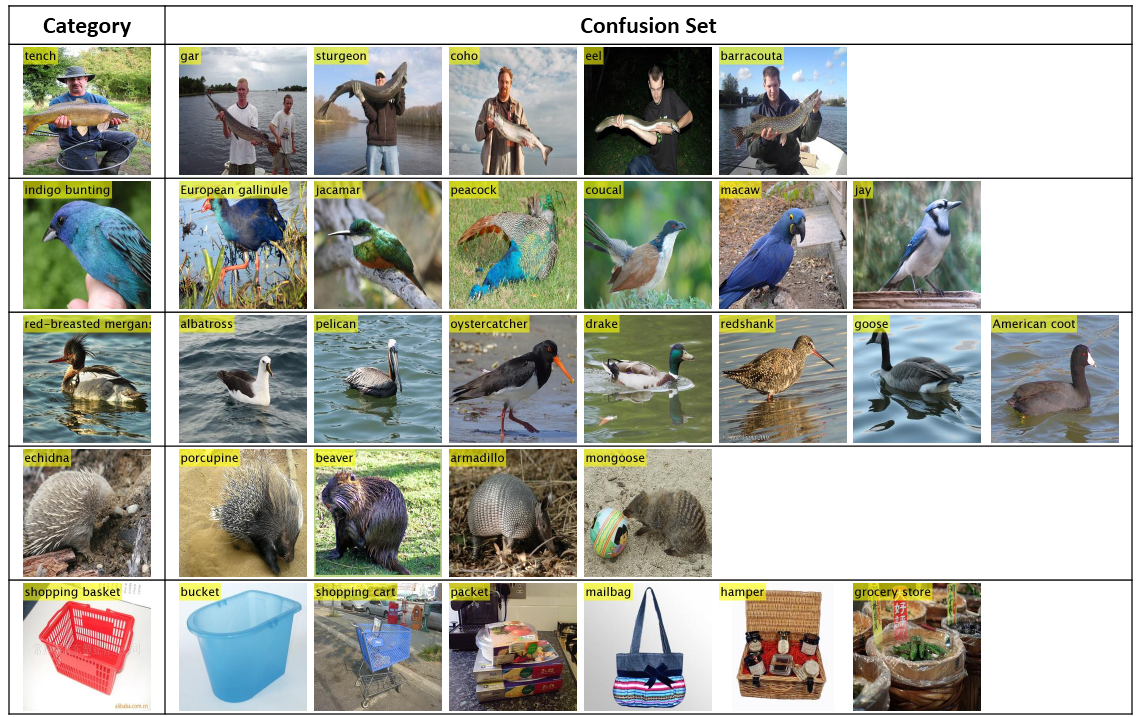} 
\caption{Example images of some categories and their corresponding confusion sets obtained by using softmax confusion matrix of the basic CNN 
on the ILSVRC 2015 dataset. Note that different categories may have different numbers of confusion classes.}
\label{fig:confusion_set} 
\end{figure*} 

Learning a CNN tree $T=(V,E)$ refers to 
the learning of tree structure as well as the specific CNN model at each node. 
For each non-leaf node, 
we propose a softmax confusion matrix to robustly estimate the
confusion sets of classes at this node 
and develop a heuristic algorithm to efficiently merge them into several confusion supersets.
For each non-root node, we show how to learn a specific CNN model to 
distinguish the classes at this node.

\subsubsection{Confusion Set Estimation\label{sec:confuse_estimation}}

Given a non-leaf $v$ of a CNN tree, 
we need to estimate the confusion set of each $c_i\in\mathcal{P}(v)$.
We perform the estimation based on the confusion matrix.
The confusion matrix is a common way to evaluate the performance of a multiclass classification algorithm, which plots the actual class label against the predicted class label.
Specifically, the confusion matrix for an $n$-class problem is an $n\times n$ matrix $H$,
where the ${ij}^{th}$ entry $h_{ij}$ represents the percentage of test instances of class $c_i$ that were 
predicted to belong to class $c_j$.
Some researchers have already proposed to explore the relationships among classes based on the confusion matrix ~\cite{godbole_scaling_2002,godbole_discriminative_2004,bengio_label_2010, griffin_learning_2008}.
The performance of their algorithms mainly depends on the accuracy of 
the confusion matrix computed by using a held-out validation dataset.
However, for the large-scale multiclass classification problem, it is impractical to 
collect a large enough validation dataset to compute the confusion matrix for accurately measuring
the inter-class relationships. 
For example, the ILSVRC 2015 dataset~\cite{ILSVRC_2015} 
provides a validation dataset with only 50 samples for each class.
As there are a total of 1000 different classes, the confusion matrix computed on such a small validation 
dataset cannot sufficiently capture the relationships among all these classes.
For rich training data, it is not appropriate to use them to 
directly compute the confusion matrix based on the classification results
since the model will fit 
the training data and can not produce 
a meaningfull confusion matrix on them.

To deal with the above problems, we propose a variant of confusion matrix which could take the advantage of
rich training data.
We observe that the output of the softmax layer in CNN represents the predicted 
confidence scores of the input instances against all the classes.
Such scores indeed capture the confusion information between classes and are more robust to overfitting.
Thus, by summing over the predicted confidence scores of all the training data, it is reasonable 
to obtain a robust estimation of the confusion information among all classes.
We refer to such a variant as softmax confusion matrix.

For the non-leaf node $v$, 
suppose there are $n_v$ classes (\ie $\left|\mathcal{P}(v)\right| = n_v$)
and $t_i$ training examples for class $c_i$.
Let $x_{ik}$ be the $k^{th}$ training example of $c_i$ and $\mathbf{p}_{ik}$ be
$n_v$-dimensional output vector of the softmax layer of the current CNN model $M(v)$ for $x_{ik}$.
Then, the softmax confusion matrix $\tilde H$ can be computed as
\begin{equation}
\tilde H = \left[ {\begin{array}{*{20}{c}}
\mathbf{\tilde h}^T_{1,\mathord{\cdot}}\\
\mathbf{\tilde h}^T_{2,\mathord{\cdot}}\\
 \vdots \\
\mathbf{\tilde h}^T_{n_v,\mathord{\cdot}}
\end{array}} \right]
\label{eq:softmax_cm}
\end{equation}
where
\begin{equation}
\mathbf{\tilde h}_{i,\cdot} = \frac{1}{t_i}\sum\limits_{k = 1}^{t_i}{\mathbf{p}_{ik}}
\end{equation}

With the above definition, 
the ${ij}^{th}$ entry $\tilde{h}_{ij}$ of $\tilde H$ 
measures how likely the examples of $c_i$ will be classified as $c_j$ on the current model $M(v)$.
Considering the $j^{th}$ column $\mathbf{\tilde h}_{\cdot,j}$,   
it consists of all the scores that measures how likely the examples of their corresponding classes 
will be classified as $c_j$ on $M(v)$.
Given an appropriate threshold $\alpha$,
a set $\mathcal{S}_j$ can be derived from $\mathbf{\tilde h}_{\cdot,j}$ 
by only reserving the class labels of entries that are larger than $\alpha$.
As $\mathcal{S}_j$ consists of classes that could be confused by $c_j$,
\ie the most possible classes that could be predicted as $c_j$, 
we regard it as an estimation of the confusion set of $c_j$.
Fig.~\ref{fig:confusion_set} shows example images of 
some classes in the ILSVRC 2015 training dataset~\cite{ILSVRC_2015} and their confusion sets estimated by using the softmax confusion matrix of the AlexNet~\cite{Alex_imagenet_2012}.
It can be observed that each class has some similar
aspects (\eg appearance, shape, size \etc) with its confusion classes.
More specific features are needed to distinguish them from their confusion classes.

\subsubsection{Confusion Set Packing\label{sec:confuse_packing}}

Given a non-leaf node $v$ and the 
confusion set for each class in $\mathcal{P}(v)$ estimated 
by using softmax confusion matrix of $M(v)$, 
we will merge these confusion sets into several confusion supersets.
We may sacrifice accuracy because now our
specific CNN model deals with more categories, but at
the same time, we gain on reducing the number of CNN
models to train.
We aim to have as few confusion supersets  as possible, with the constraint
that the error rate of each superset's CNN model should be low.
However, we cannot try all the combinations to find the one which results
in the smallest number of confusion supersets under this constraint.
Instead, we change the constraint as limiting the number of classes in each confusion superset.
It is based on the assumption that the performance of a CNN is 
highly relevant to the number of classes it aims to distinguish. 

\begin{algorithm}
\begin{algorithmic}[1]
\Require the size limit $L$, the confusion sets $\mathcal{S}_i$ for all $c_i, i\in\{1,\dots,n\}$.
\Ensure  several confusion supersets $\mathcal{Q}_i, i\in\{1,\dots,N\}$
\For{$i=1,\dots, N$}
\State $\mathcal{B}_i \gets \mathcal{S}_i$
\EndFor
\While{true}
\State Find the 'optimal' index pair $i,j$ of current step as $\underset{i,j}{\arg\max}\ {\left|{\mathcal{B}_i \cap \mathcal{B}_j}\right|}, \forall \mathcal{B}_i, \mathcal{B}_j \neq \emptyset \wedge \left|{\mathcal{B}_i \cup \mathcal{B}_j}\right| \leq L$. 
\If{No such index pair can be found}
\Break
\Else
\If{More than one index pair are found}
\State Select only one index pair $i,j$ randomly.
\EndIf
\State $\mathcal{B}_{i} \gets \mathcal{B}_{i} \cup \mathcal{B}_{j}$
\State $\mathcal{B}_{j} \gets \emptyset$
\EndIf
\EndWhile
\State $i\gets 1$
\For{$k=1,\dots, n$}
\If{$\mathcal{B}_k\neq \emptyset$}
\State $\mathcal{Q}_i \gets \mathcal{B}_k$
\State $i\gets i+1$
\EndIf
\EndFor
\end{algorithmic}
\caption{\emph{The heuristic algorithm of confusion set packing}}
\label{alg:greedy}
\end{algorithm}

Let $\mathcal{Q}$ denote the confusion superset.
At a given limit $L$ of $\left|\mathcal{Q}\right|$, 
the optimal solution is the one that minimizes the number of generated $\mathcal{Q}$.
This problem is similar to the well-studied bin packing problem which aims to
pack objects of different volumes into a finite number of 
bins with limit volume in a way that minimizes the number of bins used.
Given $n$ candidate supersets $\mathcal{B}_i, i\in\{1,\dots,n\}$ with the size limit $\left|L\right|$ 
and the confusion sets $\mathcal{S}_i, i\in\{1,\dots,n\}$ for all $n$ classes,
 the confusion set packing problem can be formulated as:

Minimize
\begin{equation}
N = \sum\limits_{i = 1}^n {{y_i}} 
\label{eq:mini}
\end{equation}
subject to
\begin{equation}
\begin{aligned}
\left|\bigcup\limits_{j = 1}^n {\widehat{\mathcal{S}}_{ij}}\right| \le {y_i}L ,\forall i \in \{1,\dots,n\} \\
\sum\limits_{i = 1}^n \left|{\widehat{\mathcal{S}}_{ij}}\right| =\left|\mathcal{S}_j\right| ,\forall j \in \{ 1,\dots,n\} \\
\end{aligned}
\label{eq:constraint}
\end{equation}
where
\begin{equation}
\begin{aligned}
y_i &=
    \begin{cases}
      1, & \text{if}\ \text{$\mathcal{B}_i$ is used} \\
      0, & \text{otherwise}
    \end{cases}, 
    \forall i \in \{ 1,\dots,n\}\\    
{\widehat{\mathcal{S}}_{ij}} &=
    \begin{cases}
      \mathcal{S}_j, & \text{if}\ \text{$\mathcal{S}_j$ is packed into $\mathcal{B}_i$}\\
      \emptyset, & \text{otherwise} 
    \end{cases},
    \forall i,j \in \{1,\dots,n\}\\
\end{aligned}
\label{eq:where}
\end{equation}
Note that the first constraint in Eq.~\ref{eq:constraint} ensures the size limit of confusion supersets
while the second one ensures that one confusion set 
can be packed into one confusion superset once and only once.

Given a solution of the problem, the $N$ confusion supersets $\mathcal{Q}_k, k\in\{1,\dots,N\}$
can be found by:
\begin{equation}
\begin{aligned} 
\mathcal{Q}_k &= \{ \bigcup{\mathcal{S}_j}|\widehat{\mathcal{S}}_{{i_k}j} \neq \emptyset, j=1,2,\dots,n \},\ \forall y_{i_k}=1
\end{aligned}
\end{equation}

Unlike the traditional bin packing problem,
the cumulative size of several confusion sets can be smaller than the sum of individual confusion set sizes due to overlap, 
\ie
$\left|\bigcup\limits_{j = 1}^n {\widehat{\mathcal{S}}_{ij}}\right| \leq \sum\limits_{j=1}^n{\left|{\widehat{\mathcal{S}}_{ij}}\right|}$.
Actually, the confusion set packing problem is 
a virtual machine packing problem with general sharing model as described in ~\cite{sindelar_sharing-aware_2011} where the different virtual machines can share space when packed into a physical server.
Finding an optimal solution of such a problem is very difficult since it has an NP-hard computational complexity.
~\cite{sindelar_sharing-aware_2011} shows that it is even infeasible to get a well approximated solution.

Here we propose a heuristic algorithm based on the intuition 
that confusion sets with more overlap should be packed 
together in higher priority in order to optimize the benefits of inter-sets sharing.
The algorithm initializes $n$ confusion supersets by allocating one confusion set to one superset respectively.
At each step, the algorithm greedily searches 
the superset pair who has the maximal overlap 
among all possible superset pairs whose union set do not 
exceed the confusion superset size limit, and then merge them together.
The detailed algorithm is described in Algorithm~\ref{alg:greedy}.

\subsubsection{Learning Specific CNN Model\label{sec:fine_grain}}

Given a non-root node $v$ and a class subset $\mathcal{P}(v)$, 
we will learn a specific CNN model for this subset only.

Let $v'$ be the parent node of $v$, $\left|\mathcal{P}(v)\right| = n_v$.
We learn $M(v)$ by fine-tuning the CNN model of $v'$, \ie $M(v')$, on $P(v)$.
In this way, we can still leverage visual knowledge from a larger number of categories.
Specifically, we follow the procedure described in~\cite{Girshick_rich_2014}.
First, we remove the last fully-connected layer of $M(v')$.
Next, we append a new randomly initialized fully-connected layer with $n_v$ output units.
Then, we run stochastic gradient descent (SGD) to learn the new CNN model
on training data of $\mathcal{P}(v)$.
As  CNN could simultaneously learn 
convolutional features and classification weights, 
the fine-tuning procedure should 
encode more discriminative
visual patterns which are more powerful to distinguish classes in $P(v)$.

\subsubsection{Summary\label{sec:sumary}}

We summarize the algorithm for learning the CNN tree in Algorithm~\ref{alg:cnn_tree_train}.
The algorithm starts by initializing the 
root node $\tilde{v}$ with the whole class set $\tilde{\mathcal{C}}$ and a given basic CNN model $\tilde{M}$.
Then a recursive procedure is performed to grow the tree in a top-down breadth-first manner 
until it reaches a given maximal depth.
With the growing of the tree, we can learn specific CNN models, 
each of which aims to learn fine-grained features for a number of categories only.

\begin{algorithm}
\begin{algorithmic}[1]
\Require The basic CNN model $\tilde{M}$ trained on $\tilde{C}$, 
the maximal tree depth $D$, 
the size limits $L_d$ of confusion superset 
in depths $d\in\{0,1,\dots,D-1\}$.
\Ensure The CNN tree.

\Function{Train}{$v$, $d$}

\If{$d\neq 0$} 
\State Fine-tune the CNN model $M(v)$ on $P(v)$
\EndIf
\If {$d == D$}	
\Return
\EndIf
\State Compute softmax confusion matrix $\tilde H$ of $M(v)$ by using Eq.~\ref{eq:softmax_cm}
\State Get $S_i$ of each class $c_i\in\mathcal{P}(v)$ from $\tilde H$
\State Run Algorithm~\ref{alg:greedy} to get the confusion supersets $\mathcal{Q}_k, k=\{1,\dots, N\}$ at the size limit $L_d$.
\For{each $k=\{1,\dots, N\}$}
\State Generate a node $u_k$ as a new child node of $v$.
\State $\mathcal{P}(u_k) \gets \mathcal{Q}_k$
\State \Call{Train}{$u_k$, $d+1$}
\EndFor
\EndFunction
\\
\State $\mathcal{P}(\tilde{v}) \gets \tilde{C}$
\State $M(\tilde{v}) \gets \tilde{M}$
\State \Call{Train}{$\tilde{v}$, $0$}
\end{algorithmic}
\caption{\emph{The training algorithm of hierarchically specific CNN}}
\label{alg:cnn_tree_train}
\end{algorithm}

\section{Experiments\label{sec:exp}}

\subsection{Dataset and The Evaluation Criteria }

\begin{table}[]
\scriptsize
\centering
\begin{tabular}{|c|c|c|c|c|}
\hline
             & \multicolumn{4}{c|}{$L_1$}                \\ \hline
             & 50      & 100     & 150     & 200     \\ \hline
Top-1 errors & 40.66\% & 40.68\% & 40.79\% & 40.86\% \\ \hline
Top-5 errors & 18.78\% & 18.58\% & 18.49\% & 18.53\% \\ \hline
\end{tabular}
\caption{Top-1 and top-5 error rates of different values of $L_1$ when using 1 AlexNet as basic model}
\label{tb:param_L1}
\end{table}

\begin{table}[]
\scriptsize
\centering
\begin{tabular}{|c|c|c|c|}
\hline
             & \multicolumn{3}{c|}{$L_2$}                \\ \hline
             & 30      & 50     & 70          \\ \hline
Top-1 errors & 40.40\% & 40.40\% & 40.43\%  \\ \hline
Top-5 errors & 18.56\% & 18.55\% & 18.60\%  \\ \hline
\end{tabular}
\caption{Top-1 and top-5 error rates of different values of $L_2$ when using 1 AlexNet as basic model and setting $L_1$ to 100}
\label{tb:param_L2}
\end{table}

We evaluate the proposed method on the image classification 
task with the ILSVRC 2015 dataset~\cite{ILSVRC_2015}.
This dataset contains 1000 leaf-node categories selected from the Imagenet hierarchy.
There are about 1.2 million images for training and 50,000 images for validation, each of which
 is associated with a ground truth category.
Following the previous papers, we test our method on the validation data.
Two criteria are used to evaluate the 
algorithms: the top-1 error rate, which compares the 
first predicted class label against the ground truth, 
and the top-5 error rate, which checks whether the ground truth is 
among the top 5 predicted class labels.

We use both AlexNet~\cite{Alex_imagenet_2012} and GoogleNet~\cite{Christian_going} as our basic CNN models.
To make a fair comparison, we use the Caffe~\cite{jia2014caffe} implementation and download the pre-trained models for the ILSVRC 2015 dataset.

The parameters for building a CNN tree are:
$(1)$ the maximal tree depth $D$, 
and $(2)$ the size limit $L_d$ of confusion superset at level $d\in\{0,1, \dots, D-1\}$.
We tested the effects of using different parameters on AlexNet. 
For $L_1$ with 50, 100, 150 and 200, the top-1(top-5) errors are 
shown in Table~\ref{tb:param_L1}.
It can be found that different settings lead to similar performance.
By setting $L_1$ as 100, for $L_2$ with 30, 50 and 70, the top-1(top-5) errors 
are shown in Table~\ref{tb:param_L2}. The performance also are similar.
Actually, We do not optimize the parameters on the validation set. 
We empirically selected the parameters as shown in Table~\ref{tb:parameters} since they don't affect the performance much.
Note that we select $D$ as 2 and 1 for AlexNet and GoogleNet respectively since a deeper tree requires to train more CNNs 
but it only leads to a small increase in performance.

\begin{table}[]
\centering
\begin{tabular}{|c|c|c|c|c|c|}
\hline
\multicolumn{1}{|l|}{} & \multicolumn{3}{c|}{AlexNet}                                               & \multicolumn{2}{l|}{GoogleNet}                   \\ \hline
Parameter              & \multicolumn{1}{c|}{$D$} & \multicolumn{1}{c|}{ $L_1$} & \multicolumn{1}{c|}{ $L_2$} & \multicolumn{1}{c|}{$D$} & \multicolumn{1}{c|}{ $L_1$} \\ \hline
Value                  & 2                      & 100                     & 50                      & 1                      & 100                     \\ \hline
\end{tabular}
\caption{The selected parameters for our CNN tree.}
\label{tb:parameters}
\end{table}

\subsection{Results}

We firstly compare our CNN tree at different depths with the basic CNN model.
We use both AlexNet and GoogleNet as our basic models since they are representative architecture which are widely used.
We denote the CNN tree with depth $D$ as $T_D$ (Note that $T_0$ is the basic CNN model).
The results for AlexNet and GoogleNet on ILSVRC 2015 are summarized in Table~\ref{tb:alex_net} and Table~\ref{tb:google_net}
respectively (the column that the number of basic model is 1).
For AlexNet, $T_1$ decreases top-1 error rate by 2.41\% and top-5 error rate by 1.43\% when compared with $T_0$.
When we increase the depth to 2, $T_2$ decreases top-1 error rate by 2.69\% and top-5 error rate by 1.49\%.
For GoogleNet, $T_1$ decreases the top-1 error rate by 4.38\% and the top-5 error rate by 1.91\%.
The improvements are significant especially on GoogleNet since the task is to classify 50,000 examples into 1,000 different classes.
The results demonstrate that the proposed CNN tree can enhance the discriminability of the basic CNN.
The reason can be explained by the fact that the specific CNN models could learn 
more discriminative features to distinguish the confused classes.
One may argue that the performance of a standard CNN can also be improved by
increasing its the depth and width.
By comparing the results on AlexNet and GoogleNet, we note that our method is even better at improving the performance of a strong basic model, \eg GoogleNet. 
This may due to that the confusing sets estimated by a strong basic model are more 
accurate than those estimated by a weak basic model.
It means that our method can enjoy the benefits of optimizing the single basic model. In practice, we can combine the CNN tree with the optimized basic model to further boost the performance.

As the performance of a single model can always be improved
by averaging multiple models, we then investigate
whether the proposed method can also improve multiple basic
models. We first train several basic CNN
models by different random initializations and then train a CNN tree
for each basic model. The final prediction is obtained by
averaging the predictions of CNN trees of different basic
models. 
The results for AlexNet and GoogleNext are also shown in Table~\ref{tb:alex_net} and Table~\ref{tb:google_net} respectively.
As can be observed, the performance could be further improved by applying CNN trees to multiple basic
models.This may be because the confusion
sets estimated by multiple basic models are more robust
than those estimated by a single basic model, which helps to correct more misclassified samples. 
It also
indicates that the proposed method provides values beyond
model averaging.
For example, averaging 6 basic models of AlexNet decreases
 top-1 (top-5) errors from 43.09\% (20.04\%) to 39.63\%
(17.43\%) while applying our CNN trees on these 6 basic
models further decreases the top-1 (top-5) errors to 37.19\%
(16.23\%). On the other hand, averaging 6 basic models of GoogleNet decreases
 top-1 (top-5) errors from 32.75\% (12.00\%) to 29.56\%
(10.08\%) while applying our CNN trees further decreases top-1 (top-5)
errors to 25.15\% (8.12\%).

\begin{table*}[]
\scriptsize
\centering
\begin{tabular}{|c|c|l|l|l|l|l|l|}
\hline
\multicolumn{2}{|l|}{}                                                                             & \multicolumn{6}{c|}{AlexNet}                                                                                                                        \\ \hline
\multicolumn{1}{|l|}{}                                                            & \#Basic Models & \multicolumn{1}{c|}{1} & \multicolumn{1}{c|}{2} & \multicolumn{1}{c|}{3} & \multicolumn{1}{c|}{4} & \multicolumn{1}{c|}{5} & \multicolumn{1}{c|}{6} \\ \hline
\multirow{3}{*}{\begin{tabular}[c]{@{}c@{}}Top-1 \\ errors\end{tabular}} &  $T_0$             & 43.09\%                & 41.28\%                & 40.41\%                & 40.21\%                & 39.82\%                & 39.63\%                \\ \cline{2-8} 
                                                                                  &  $T_1$             & 40.68\%(-2.41\%)       & 38.95\%(-2.33\%)       & 38.07\%(-2.34\%)       & 37.80\%(-2.41\%)       & 37.49\%(-2.33\%)       & 37.39\%(-2.24\%)       \\ \cline{2-8} 
                                                                                  &  $T_2$             & 40.40\%(-2.69\%)       & 38.60\%(-2.68\%)       & 37.84\%(-2.57\%)       & 37.62\%(-2.59\%)       & 37.33\%(-2.49\%)       & 37.19\%(-2.44\%)       \\ \hline
\multirow{3}{*}{\begin{tabular}[c]{@{}c@{}}Top-5\\ errors\end{tabular}}    &  $T_0$             & 20.04\%                & 18.53\%                & 18.03\%                & 17.72\%                & 17.52\%                & 17.43\%                \\ \cline{2-8} 
                                                                                  &  $T_1$             & 18.58\%(-1.46\%)              
& 17.52\%(-1.01\%)       & 16.93\%(-1.10\%)       & 16.59\%(-1.13\%)       & 16.39\%(-1.13\%)       & 16.24\%(-1.19\%)                                                                                                                                                                          \\ \cline{2-8} 
                                                                                  &  $T_2$             & 18.55\%(-1.49\%)       & 17.39\%(-1.14\%)       & 16.81\%(-1.22\%)       & 16.53\%(-1.19\%)       & 16.36\%(-1.16\%)       & 16.23\%(-1.20\%)       
                                                                                  \\ \hline
\end{tabular}
\caption{Top-1 and top-5 error rates of different number of basic models on validation data of ILSVRC 2015 by using AlexNet. 
The basic Model is denoted as $T_0$.
The CNN trees with depth $D=1$ and $D=2$ are denoted as $T_1$ and $T_2$ respectively. The scores in brackets are the decreased errors compared to $T_0$.}
\label{tb:alex_net}
\end{table*}

\begin{table*}[]
\scriptsize
\centering
\begin{tabular}{|c|c|l|l|l|l|l|l|}
\hline
\multicolumn{2}{|l|}{}                                                                             & \multicolumn{6}{c|}{GoogleNet}                                                                                                                      \\ \hline
\multicolumn{1}{|l|}{}                                                            & \#Basic Models & \multicolumn{1}{c|}{1} & \multicolumn{1}{c|}{2} & \multicolumn{1}{c|}{3} & \multicolumn{1}{c|}{4} & \multicolumn{1}{c|}{5} & \multicolumn{1}{c|}{6} \\ \hline
\multirow{2}{*}{\begin{tabular}[c]{@{}c@{}}Top-1\\  errors\end{tabular}} & T0             & 32.75\%                & 30.96\%                & 30.27\%                & 29.89\%                & 29.72\%                & 29.56\%                \\ \cline{2-8} 
                                                                                  & T1             & 28.37\%(-4.38\%)       & 26.51\%(-4.45\%)       & 25.99\%(-4.28\%)       & 25.57\%(-4.32\%)       & 25.4\%(-4.32\%)        & 25.15\%(-4.41\%)       \\ \hline
\multirow{2}{*}{\begin{tabular}[c]{@{}c@{}}Top-5\\ erros\end{tabular}}    & T0             & 12.00\%                & 10.89                  & 10.53\%                & 10.32\%                & 10.17\%                & 10.08\%                \\ \cline{2-8} 
                                                                                  & T1             & 10.09\%(-1.91\%)       & ~~8.98\%(-1.91\%)        & ~~8.68\%(-1.85\%)        & ~~8.33\%(-1.99\%)        & ~~8.23\%(-1.94\%)        & ~~8.12\%(-1.96\%)                 \\ \hline
\end{tabular}
\caption{Top-1 and top-5 error rates of different basic models on validation data of ILSVRC 2015 by using GoogleNet. 
The basic Model is denoted as $T_0$
The CNN tree with depth $D=1$ is denoted as $T_1$. The scores in brackets are the decreased errors compared to $T_0$.}
\label{tb:google_net}
\end{table*}

\begin{table*}[]
\scriptsize
\centering
\begin{tabular}{|c|c|c|c|c|}
\hline
           & \multicolumn{2}{c|}{AlexNet}    & \multicolumn{2}{c|}{GoogleNet} \\ \hline
BP Layers  & Top-1 errors   & Top-5 errors   & Top-1 errors   & Top-5 errors  \\ \hline
All Layers & 40.68\%(-2.41\%) & 18.58\%(-1.46\%) &      28.37\%(-4.38\%)          &       10.09\%(-1.91\%)      \\ \hline
Last Layer & 42.96\%(-0.13\%) & 20.14\%(+0.10\%) &      32.16\%(-0.59\%)          &       11.75\%(-0.25\%)      \\ \hline
\end{tabular}
\caption{Comparison between different learning strategies.
'All layers' means we perform back-propagation for all layers of the CNN and 'Last layer' means 
we perform back-propagation only for the last fully-connected layer.}
\label{tb:bp_test}
\end{table*}

\begin{figure}[!htb]
 \centering
\subfigure[AlexNet top-1 error rate ]{       
        \includegraphics[width=0.40\textwidth]{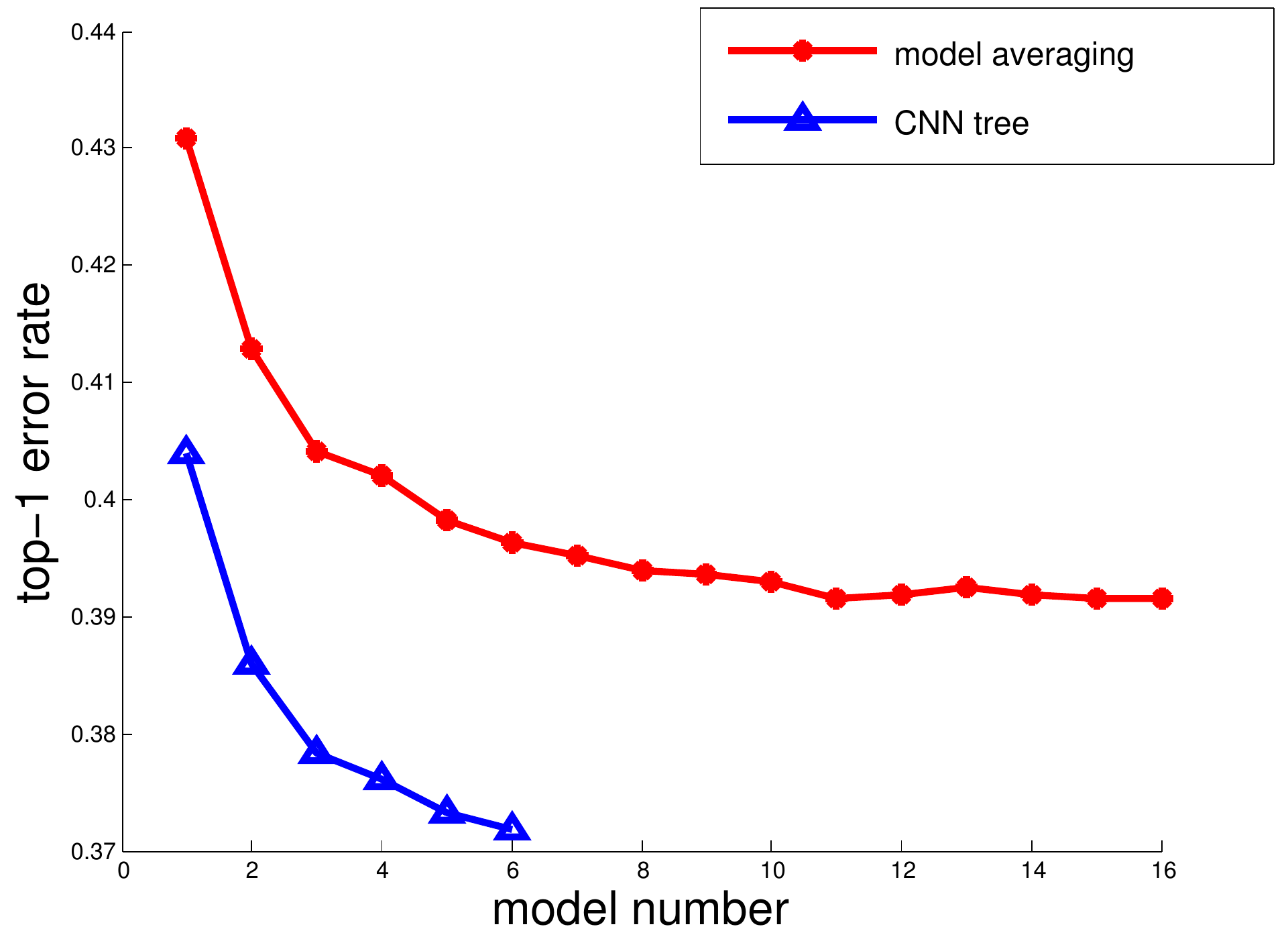}}
\subfigure[ALexNet top-5 error rate]{
        \includegraphics[width=0.40\textwidth]{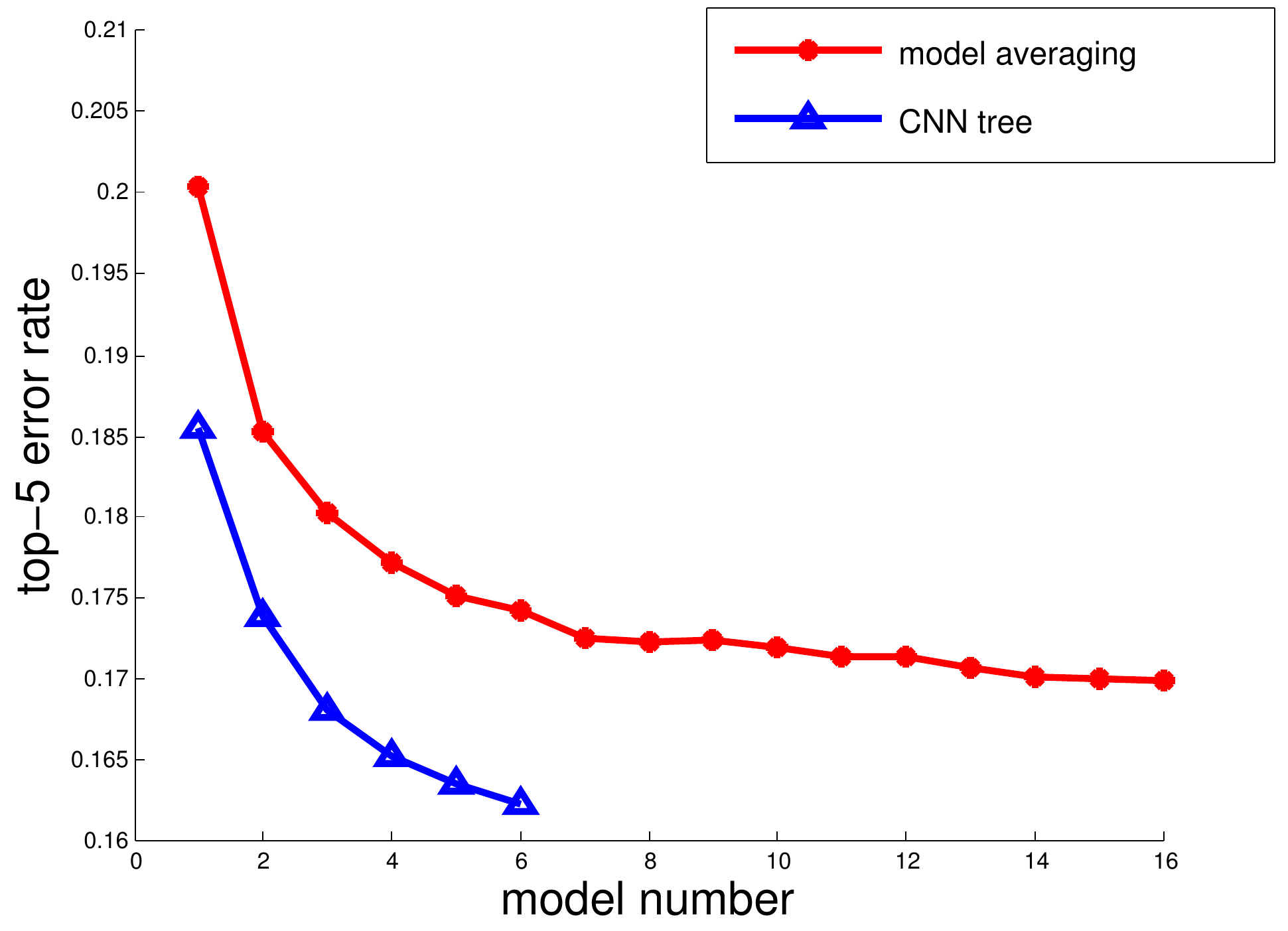}}
\subfigure[GoogleNet top-1 error rate ]{       
        \includegraphics[width=0.40\textwidth]{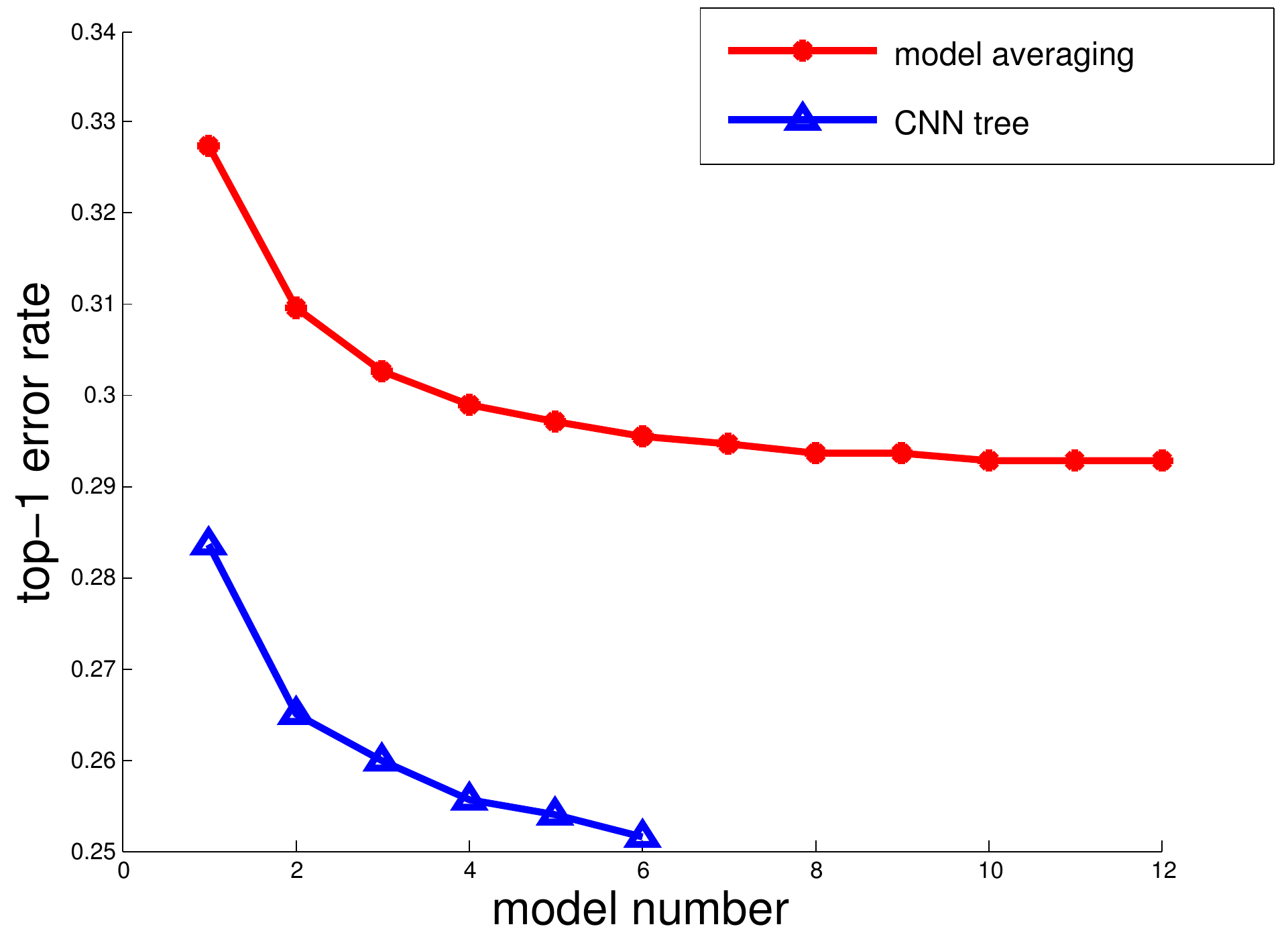}}
\subfigure[GoogleNet top-5 error rate]{
        \includegraphics[width=0.40\textwidth]{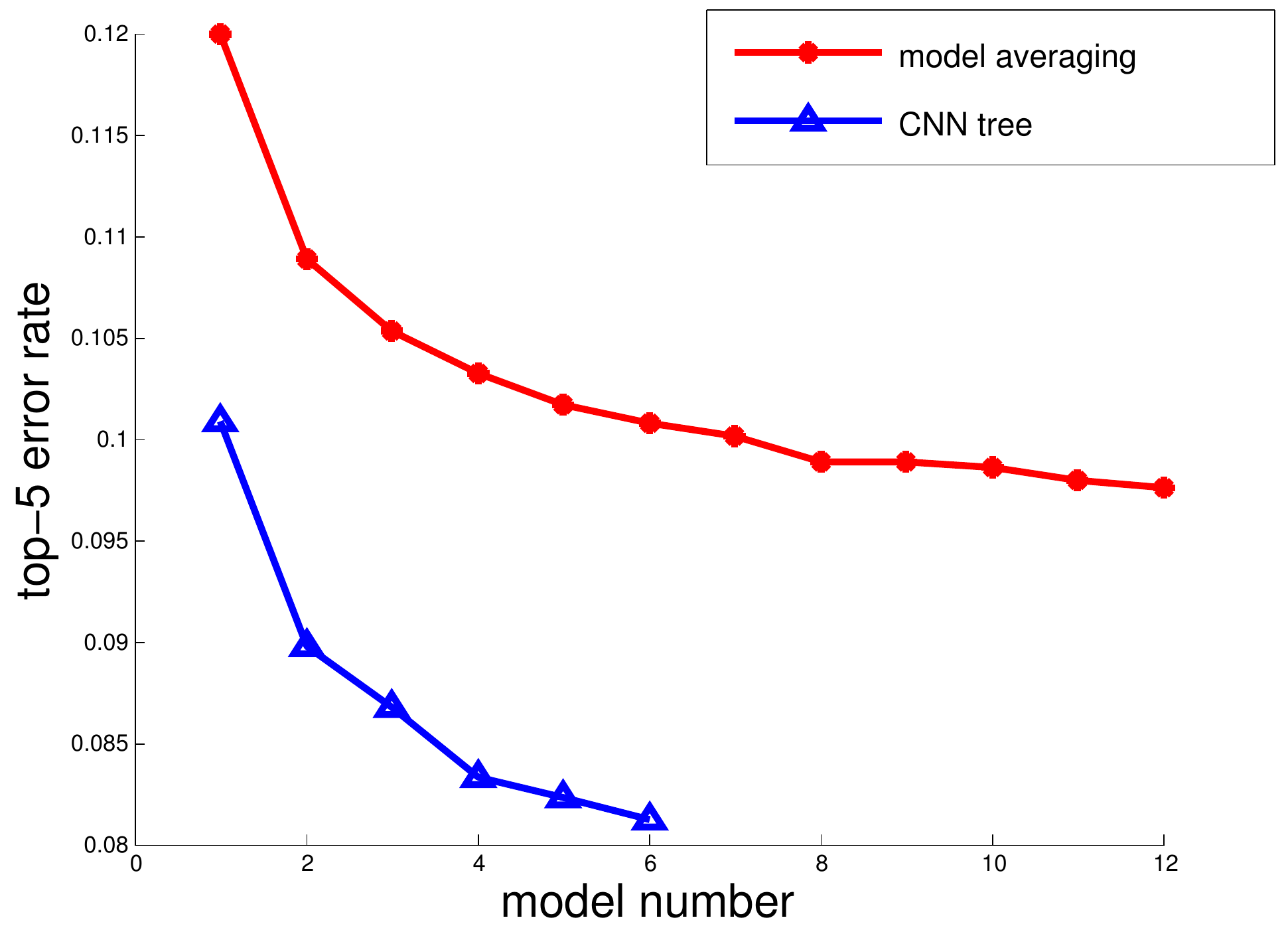}}        
\caption{ The decrement of top-1 and top-5 error rates as a function of model number.\label{fig:average} }
\end{figure}

To show the limitation of model averaging,
we present the error decrement with respect to the model number in model averaging in Fig~\ref{fig:average}. 
For AlexNet, averaging 16 models decreases top-1(top-5) errors to 39.15(\%16.99\%). 
For GoogleNet, averaging 12 models decreases top-1(top-5) errors to 29.29\%(9.76\%). 
As can be observed, model averaging can only decrease the error rate to some extent, and no further decrement can be obtained by using more models. 
However, by combining our methods with model averaging, 
\ie applying the proposed CNN tree on the averaged basic models, the error rate can be significantly decreased.
This sufficiently demonstrates the extra value of our 
method over model averaging.

We further conduct an experiment to show our method 
can learn more powerful feature representation at lower layers in the network.
We try another learning strategy:
For each class subset in the non-root node, 
we fix the weights of all the layers except for the last fully-connected layer, 
and then fine-tune the CNN model for this class subset.
With this setting, we actually train a multiclass logistic regression classifier 
for the class subset without learning features.
We compare this learning strategy with $T_1$.
As shown in Table~\ref{tb:bp_test}, 
for AlexNet, fine-tuning without feature learning
decreases the top-1 error rate by 0.13\% and even increases the top-5 error rate by 0.1\%. 
For GoogleNet, fine-tuning without feature learning
decreases the top-1 and top-5 error rate by  0.59\% and 0.25\% respectively.
This strategy performs much worse than fine-tuning the specific CNN models with feature learning.
This demonstrates that the proposed CNN tree enhances the discriminability 
by progressively learning fine-grained features for each class subset.

\begin{figure*} 
\centering\includegraphics[width=1.0\textwidth]{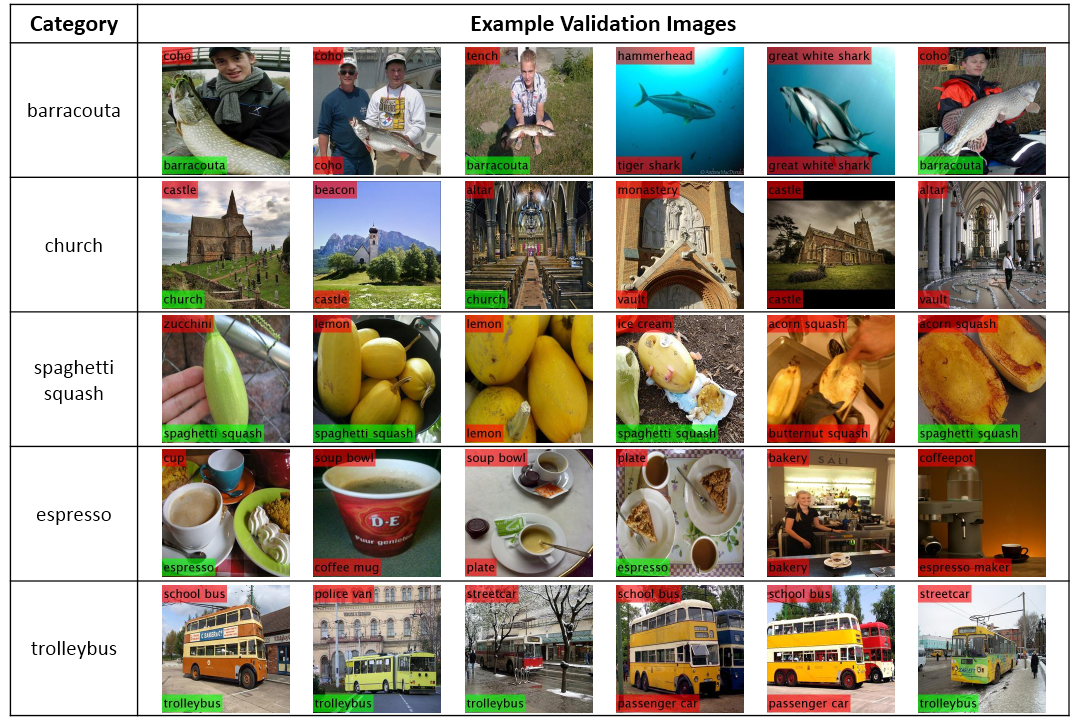} 
\caption{ Top-1 predictions of some categories on validation data.
Each image is marked with two labels, where the top label is predicted by the basic CNN model
and the bottom label is predicted by using the proposed CNN tree.
The red (green) color denotes that the prediction is wrong (right) compared to the ground truth.
}
\label{fig:misclassify}
\end{figure*}

Fig.~\ref{fig:misclassify} gives the top-1 predictions of
some categories on validation data by using AlexNet.
Each image is marked with two labels, where the top label 
is predicted by the basic CNN model
and the bottom label is predicted by using the proposed CNN tree.
The red color denotes a right prediction while the green color denotes a wrong prediction.
As can be seen, our method could correct the prediction of some examples 
that are misclassified by the basic CNN model. 
The reason is that our method could refine the class labels predicted by the basic model via progressively learning more specific models on their confusion sets.

\subsection{Computational Complexity Analysis\label{sec:complexity}}
With the parameter settings in Table~\ref{tb:parameters}, 
for AlexNet, we actually fine-tune 55 specific CNN models at level 1, each of which contains no more than 100 classes,
and 151 specific CNN models at level 2, each of which contains no more than 50 classes.
For GoogleNet, we fine-tune 61 specific CNN models 
at level 1, each of which also contains no more than 100 classes

To fine-tune these models, 
we start SGD from a learning rate of 0.001 (1/10 of the initial pre-training rate) 
and decrease the learning rate by a factor 
of 10 when the accuracy on validation data stops improving.
For AlexNet, the learning is stopped after about 15,000 iterations for each CNN model in level 1 and about 
5,000 iterations in level 2.
Thus, learning such a CNN tree requires a total of about 1.5 million iterations, 
which is 5 times of learning the basic model (310k iterations).
In experiments, we train the proposed CNN tree in parallel on 12 Tesla K40 GPUs.
It takes about 1.5 days to finish the training procedure.
For testing, our method increases the time by 2 times as each test examples need to be evaluated by 3 CNNs.
For GoogleNet, the learning is stopped after about 80,000 iterations for each CNN model in level 1. 
Thus it requires a total of about 4.8 million iterations, which is 2 times of learning
the basic model (2.4 million iterations).
It takes about 2 days to finish the training procedure on the same platform. The testing time is increased by 1 time
as the tree depth is 1.

\section{Conclusion\label{sec:conclusion}}
In this paper, a CNN tree is proposed to progressively 
learn fine-grained features to distinguish a subset of classes 
which are confused by the basic CNN model.
Such features are expected to be more discriminative, 
compared to features learned for all the classes.
Thus, test examples that are misclassified by the basic CNN model might
be correctly classified by the specific CNN model in the bottom layer.
To learn the tree structure as well as the fine-grained features,
a new learning algorithm is proposed to grow
the tree in a top-down breadth-first manner.
Experiments on large-scale image classification tasks
by using both AlexNet and GoogleNet
demonstrate that our method could enhance the discriminability of given basic CNN models.
The proposed method is fairly generic, hence it can potentially be used 
in combination with many other deep learning models.

\section*{References}

\bibliography{mybibfile}

\begin{thebibliography}{10}
\expandafter\ifx\csname url\endcsname\relax
  \def\url#1{\texttt{#1}}\fi
\expandafter\ifx\csname urlprefix\endcsname\relax\def\urlprefix{URL }\fi
\expandafter\ifx\csname href\endcsname\relax
  \def\href#1#2{#2} \def\path#1{#1}\fi

\bibitem{LeCun_back_1989}
Y.~LeCun, B.~Boser, J.~S. Denker, D.~Henderson, R.~E. Howard, W.~Hubbard, L.~D.
  Jackel, Backpropagation applied to handwritten zip code recognition, in:
  Neural Comput., Vol.~1, 1989, pp. 541--551.

\bibitem{lecun1998gradient}
Y.~LeCun, L.~Bottou, Y.~Bengio, P.~Haffner, Gradient-based learning applied to
  document recognition, Proceedings of the IEEE 86~(11) (1998) 2278--2324.

\bibitem{Alex_imagenet_2012}
A.~Krizhevsky, I.~Sutskever, G.~E. Hinton, Imagenet classification with deep
  convolutional neural networks, in: Advances in Neural Information Processing
  Systems 25, 2012, pp. 1097--1105.

\bibitem{Zeiler_visual_2014}
M.~Zeiler, R.~Fergus, Visualizing and understanding convolutional networks, in:
  Computer Vision – ECCV 2014, 2014, pp. 818--833.

\bibitem{He_delving_2015}
K.~He, X.~Zhang, S.~Ren, J.~Sun, \href{http://arxiv.org/abs/1502.01852}{Delving
  deep into rectifiers: Surpassing human-level performance on imagenet
  classification}, CoRR abs/1502.01852.
\newline\urlprefix\url{http://arxiv.org/abs/1502.01852}

\bibitem{Girshick_rich_2014}
R.~Girshick, J.~Donahue, T.~Darrell, J.~Malik, Rich feature hierarchies for
  accurate object detection and semantic segmentation, in: Computer Vision and
  Pattern Recognition (CVPR), 2014 IEEE Conference on, 2014, pp. 580--587.

\bibitem{Girshick_fast_2015}
R.~Girshick, Fast r-cnn, arXiv preprint arXiv:1504.08083.

\bibitem{Pierre_overfeat_2013}
P.~Sermanet, D.~Eigen, X.~Zhang, M.~Mathieu, R.~Fergus, Y.~LeCun,
  \href{http://arxiv.org/abs/1312.6229}{Overfeat: Integrated recognition,
  localization and detection using convolutional networks}, CoRR abs/1312.6229.
\newline\urlprefix\url{http://arxiv.org/abs/1312.6229}

\bibitem{ouyang2015deepid}
W.~Ouyang, X.~Wang, X.~Zeng, S.~Qiu, P.~Luo, Y.~Tian, H.~Li, S.~Yang, Z.~Wang,
  C.-C. Loy, et~al., Deepid-net: Deformable deep convolutional neural networks
  for object detection, in: Proceedings of the IEEE Conference on Computer
  Vision and Pattern Recognition, 2015, pp. 2403--2412.

\bibitem{Szegedy_scalable_2014}
C.~Szegedy, S.~Reed, D.~Erhan, D.~Anguelov,
  \href{http://arxiv.org/abs/1412.1441}{Scalable, high-quality object
  detection}, CoRR abs/1412.1441.
\newline\urlprefix\url{http://arxiv.org/abs/1412.1441}

\bibitem{erhan2014scalable}
D.~Erhan, C.~Szegedy, A.~Toshev, D.~Anguelov, Scalable object detection using
  deep neural networks, in: Computer Vision and Pattern Recognition (CVPR),
  2014 IEEE Conference on, IEEE, 2014, pp. 2155--2162.

\bibitem{li2015robust}
H.~Li, Y.~Li, F.~Porikli, Robust online visual tracking with a single
  convolutional neural network, in: Computer Vision--ACCV 2014, Springer, 2015,
  pp. 194--209.

\bibitem{wang2015video}
L.~Wang, T.~Liu, G.~Wang, K.~L. Chan, Q.~Yang, Video tracking using learned
  hierarchical features, Image Processing, IEEE Transactions on 24~(4) (2015)
  1424--1435.

\bibitem{hong2015online}
S.~Hong, T.~You, S.~Kwak, B.~Han, Online tracking by learning discriminative
  saliency map with convolutional neural network, Proceedings of the 32th
  International Conference on Machine Learning (ICML-15).

\bibitem{delakis2008text}
M.~Delakis, C.~Garcia, text detection with convolutional neural networks., in:
  VISAPP (2), 2008, pp. 290--294.

\bibitem{huang2014robust}
W.~Huang, Y.~Qiao, X.~Tang, Robust scene text detection with convolution neural
  network induced mser trees, in: Computer Vision--ECCV 2014, Springer, 2014,
  pp. 497--511.

\bibitem{Goodfellow_2013_Multi}
I.~J. Goodfellow, Y.~Bulatov, J.~Ibarz, S.~Arnoud, V.~Shet, Multi-digit number
  recognition from street view imagery using deep convolutional neural
  networks, CoRR abs/1312.6082.

\bibitem{jaderberg2014deep}
M.~Jaderberg, K.~Simonyan, A.~Vedaldi, A.~Zisserman, Deep structured output
  learning for unconstrained text recognition, arXiv preprint arXiv:1412.5903.

\bibitem{wang2012end}
T.~Wang, D.~J. Wu, A.~Coates, A.~Y. Ng, End-to-end text recognition with
  convolutional neural networks, in: Pattern Recognition (ICPR), 2012 21st
  International Conference on, IEEE, 2012, pp. 3304--3308.

\bibitem{Tian_2017}
Y.~Tian, B.~Fan, F.~Wu, L2-net: Deep learning of discriminative patch
  descriptor in euclidean space, in: Computer Vision and Pattern Recognition,
  2009. CVPR 2017. IEEE Conference on, 2017.

\bibitem{simonyan2014two}
K.~Simonyan, A.~Zisserman, Two-stream convolutional networks for action
  recognition in videos, in: Advances in Neural Information Processing Systems,
  2014, pp. 568--576.

\bibitem{karpathy2014large}
A.~Karpathy, G.~Toderici, S.~Shetty, T.~Leung, R.~Sukthankar, L.~Fei-Fei,
  Large-scale video classification with convolutional neural networks, in:
  Computer Vision and Pattern Recognition (CVPR), 2014 IEEE Conference on,
  2014, pp. 1725--1732.

\bibitem{ng2015beyond}
J.~Y.-H. Ng, M.~Hausknecht, S.~Vijayanarasimhan, O.~Vinyals, R.~Monga,
  G.~Toderici, Beyond short snippets: Deep networks for video classification,
  arXiv preprint arXiv:1503.08909.

\bibitem{toshev2014deeppose}
A.~Toshev, C.~Szegedy, Deeppose: Human pose estimation via deep neural
  networks, in: Computer Vision and Pattern Recognition (CVPR), 2014 IEEE
  Conference on, 2014, pp. 1653--1660.

\bibitem{jain2013learning}
A.~Jain, J.~Tompson, M.~Andriluka, G.~W. Taylor, C.~Bregler, Learning human
  pose estimation features with convolutional networks, arXiv preprint
  arXiv:1312.7302.

\bibitem{tompson2014joint}
J.~J. Tompson, A.~Jain, Y.~LeCun, C.~Bregler, Joint training of a convolutional
  network and a graphical model for human pose estimation, in: Advances in
  Neural Information Processing Systems, 2014, pp. 1799--1807.

\bibitem{Gong_multi_2011}
Y.~Gong, L.~Wang, R.~Guo, S.~Lazebnik, Multi-scale orderless pooling of deep
  convolutional activation features, in: Computer Vision – ECCV 2014, 2014,
  pp. 392--407.

\bibitem{Zhou_learn_2014}
B.~Zhou, A.~Lapedriza, J.~Xiao, A.~Torralba, A.~Oliva, Learning deep features
  for scene recognition using places database, in: Advances in Neural
  Information Processing Systems, 2014.

\bibitem{shuai2015integrating}
B.~Shuai, G.~Wang, Z.~Zuo, B.~Wang, L.~Zhao, Integrating parametric and
  non-parametric models for scene labeling, Computer Vision and Pattern
  Recognition (CVPR), 2014 IEEE Conference on 72 (2015) 50--8.

\bibitem{farabet2013learning}
C.~Farabet, C.~Couprie, L.~Najman, Y.~LeCun, Learning hierarchical features for
  scene labeling, Pattern Analysis and Machine Intelligence, IEEE Transactions
  on 35~(8) (2013) 1915--1929.

\bibitem{sindelar_sharing-aware_2011}
M.~Sindelar, R.~K. Sitaraman, P.~Shenoy, Sharing-aware algorithms for virtual
  machine colocation, in: Proceedings of the 23th annual {ACM} symposium on
  Parallelism in algorithms and architectures, 2011, pp. 367--378.

\bibitem{Christian_going}
C.~Szegedy, W.~Liu, Y.~Jia, P.~Sermanet, S.~Reed, D.~Anguelov, D.~Erhan,
  V.~Vanhoucke, A.~Rabinovich, Going deeper with convolutions, in: Proceedings
  of the IEEE Conference on Computer Vision and Pattern Recognition, 2015, pp.
  1--9.

\bibitem{ILSVRC_2015}
\url{http://image-net.org/challenges/LSVRC/2012/index}.

\bibitem{deng_imagenet_2009}
J.~Deng, W.~Dong, R.~Socher, L.-J. Li, K.~Li, L.~Fei-Fei, Imagenet: A
  large-scale hierarchical image database, in: Computer Vision and Pattern
  Recognition, 2009. CVPR 2009. IEEE Conference on, 2009, pp. 248--255.

\bibitem{Hinton_dropout}
G.~E. Hinton, N.~Srivastava, A.~Krizhevsky, I.~Sutskever, R.~Salakhutdinov,
  Improving neural networks by preventing co-adaptation of feature detectors,
  CoRR abs/1207.0580.

\bibitem{wan2013regularization}
L.~Wan, M.~Zeiler, S.~Zhang, Y.~L. Cun, R.~Fergus, Regularization of neural
  networks using dropconnect, in: Proceedings of the 30th International
  Conference on Machine Learning (ICML-13), 2013, pp. 1058--1066.

\bibitem{Ciresan_multi_column}
D.~Ciresan, U.~Meier, J.~Schmidhuber, Multi-column deep neural networks for
  image classification, in: Computer Vision and Pattern Recognition (CVPR),
  2012 IEEE Conference on, 2012, pp. 3642--3649.

\bibitem{goodfellow2013maxout}
I.~Goodfellow, D.~Warde-farley, M.~Mirza, A.~Courville, Y.~Bengio, Maxout
  networks, in: Proceedings of the 30th International Conference on Machine
  Learning (ICML-13), 2013, pp. 1319--1327.

\bibitem{Andrew_improve_2013}
A.~G. Howard, Some improvements on deep convolutional neural network based
  image classification, CoRR abs/1312.5402.

\bibitem{wu2015deep}
R.~Wu, S.~Yan, Y.~Shan, Q.~Dang, G.~Sun, Deep image: Scaling up image
  recognition, arXiv preprint arXiv:1501.02876.

\bibitem{Kaiming_spatial}
K.~He, X.~Zhang, S.~Ren, J.~Sun, Spatial pyramid pooling in deep convolutional
  networks for visual recognition, CoRR abs/1406.4729.

\bibitem{ioffe2015batch}
S.~Ioffe, C.~Szegedy, Batch normalization: Accelerating deep network training
  by reducing internal covariate shift, arXiv preprint arXiv:1502.03167.

\bibitem{Zeiler_visual}
M.~Zeiler, R.~Fergus, Visualizing and understanding convolutional networks, in:
  Computer Vision – ECCV 2014, Vol. 8689, 2014, pp. 818--833.

\bibitem{simonyan2014very}
K.~Simonyan, A.~Zisserman, Very deep convolutional networks for large-scale
  image recognition, arXiv preprint arXiv:1409.1556.

\bibitem{he2016deep}
K.~He, X.~Zhang, S.~Ren, J.~Sun, Deep residual learning for image recognition,
  in: Proceedings of the IEEE Conference on Computer Vision and Pattern
  Recognition, 2016, pp. 770--778.

\bibitem{griffin_learning_2008}
G.~Griffin, P.~Perona, Learning and using taxonomies for fast visual
  categorization, in: Computer Vision and Pattern Recognition, 2008. CVPR 2008.
  IEEE Conference on, 2008, pp. 1--8.

\bibitem{Deng_what_2010}
J.~Deng, A.~Berg, K.~Li, L.~Fei-Fei, What does classifying more than 10,000
  image categories tell us?, in: Computer Vision – ECCV 2010, Vol. 6315,
  2010, pp. 71--84.

\bibitem{Torralba_2008}
A.~Torralba, R.~Fergus, W.~Freeman, 80 million tiny images: A large data set
  for nonparametric object and scene recognition, Pattern Analysis and Machine
  Intelligence, IEEE Transactions on 30~(11) (2008) 1958--1970.

\bibitem{bengio_label_2010}
S.~Bengio, J.~Weston, D.~Grangier, Label embedding trees for large multi-class
  tasks, in: Advances in Neural Information Processing Systems 23, 2010, pp.
  163--171.

\bibitem{deng_fast_2011}
J.~Deng, S.~Satheesh, A.~C. Berg, F.~Li, Fast and balanced: Efficient label
  tree learning for large scale object recognition, in: NIPS, 2011, pp.
  567--575.

\bibitem{godbole_scaling_2002}
S.~Godbole, S.~Sarawagi, S.~Chakrabarti, Scaling multi-class support vector
  machines using inter-class confusion, in: Proceedings of the 18th {ACM}
  {SIGKDD} International Conference on Knowledge Discovery and Data Mining,
  2002, pp. 513--518.

\bibitem{godbole_discriminative_2004}
S.~Godbole, S.~Sarawagi, Discriminative methods for multi-labeled
  classification, in: Advances in Knowledge Discovery and Data Mining, 2004,
  pp. 22--30.

\bibitem{jia2014caffe}
Y.~Jia, E.~Shelhamer, J.~Donahue, S.~Karayev, J.~Long, R.~Girshick,
  S.~Guadarrama, T.~Darrell, Caffe: Convolutional architecture for fast feature
  embedding, arXiv preprint arXiv:1408.5093.

\end{thebibliography}

\end{document}